\documentclass[10pt,twocolumn,letterpaper]{article}

\usepackage[pagenumbers]{cvpr} 

\usepackage{graphicx}
\usepackage{amsmath}
\usepackage{amssymb}
\usepackage{booktabs}
\usepackage{color,xcolor,colortbl}
\usepackage{caption, subcaption}

\usepackage{wrapfig}

\usepackage[pagebackref,breaklinks,colorlinks]{hyperref}

\definecolor{Gray}{gray}{0.9}
\newcommand{\com}[1]{\textcolor{black}{#1}}

\newcommand\blfootnote[1]{%
  \begingroup
  \renewcommand\thefootnote{}\footnote{#1}%
  \addtocounter{footnote}{-1}%
  \endgroup
}

\usepackage[capitalize]{cleveref}
\crefname{section}{Sec.}{Secs.}
\Crefname{section}{Section}{Sections}
\Crefname{table}{Table}{Tables}
\crefname{table}{Tab.}{Tabs.}

\def\cvprPaperID{8683} 
\def\confName{CVPR}
\def\confYear{2022}

\begin{document}

\title{Hyperbolic Image Segmentation}

\author{Mina GhadimiAtigh$^{1}$\thanks{Equal contribution}, Julian Schoep$^{1}$\footnotemark[1], Erman Acar$^{2}$, Nanne van Noord$^1$, Pascal Mettes$^1$\\
$^1$University of Amsterdam, $^{2}$Leiden University, $^{2}$Vrije Universiteit Amsterdam\\
}

\twocolumn[{%
\renewcommand\twocolumn[1][]{#1}%
\maketitle
\vspace{-1cm}
\begin{center}
    \includegraphics[trim=0 190 0 0,clip=true,width=0.975\textwidth]{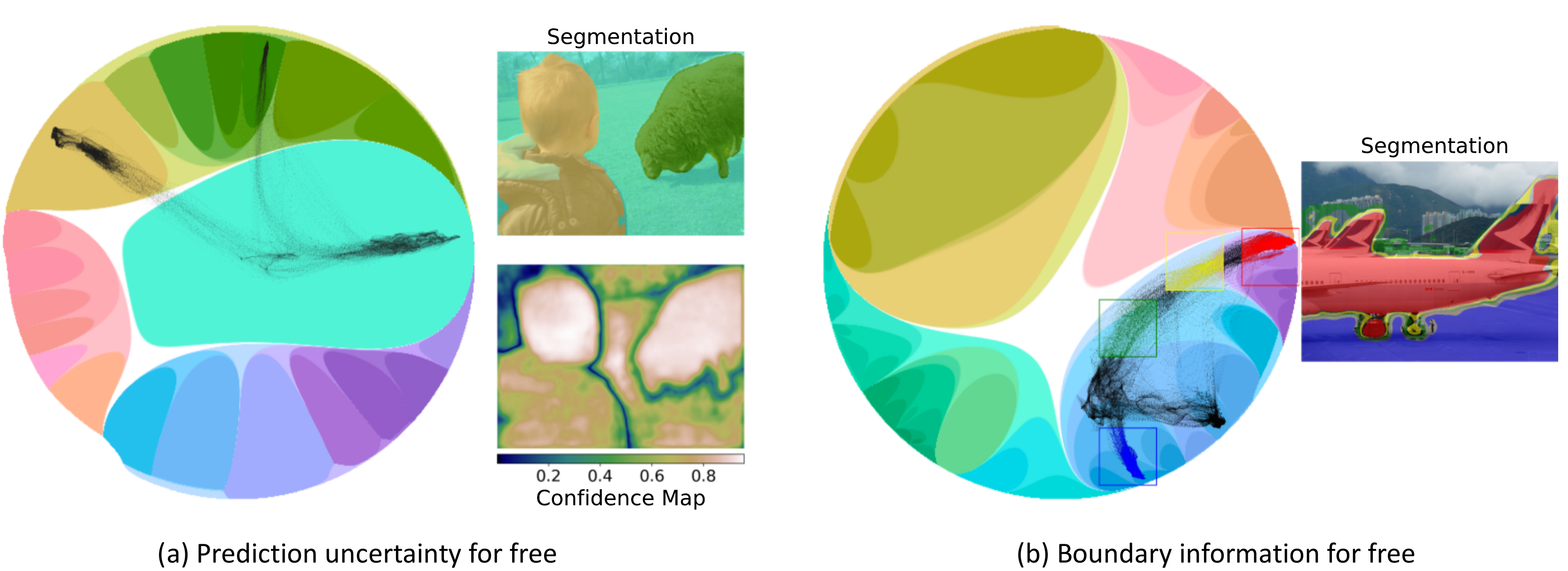}
    \captionof{figure}{\textbf{Two examples of insights that come for free with Hyperbolic Image Segmentation.} For both examples, each black dot denotes a pixel embedding in hyperbolic space. Left (Pascal VOC): next to per-pixel classification, the distance to the origin in hyperbolic space provides a free measure of uncertainty. Right (COCO-Stuff-10k): the hyperbolic positioning of pixels even allows us to pinpoint interiors and edges of objects, as indicated by the colored boxes and their corresponding pixels in the segmentation map. Other benefits of hyperbolic embeddings for segmentation include zero-label generalization and better performance in low-dimensional embedding spaces.}
    \label{fig:fig1}
\end{center}%
}]
\blfootnote{* Equal contribution}

\begin{abstract}
For image segmentation, the current standard is to perform pixel-level optimization and inference in Euclidean output embedding spaces through linear hyperplanes. In this work, we show that hyperbolic manifolds provide a valuable alternative for image segmentation and propose a tractable formulation of hierarchical pixel-level classification in hyperbolic space. Hyperbolic Image Segmentation opens up new possibilities and practical benefits for segmentation, such as uncertainty estimation and boundary information for free, zero-label generalization, and increased performance in low-dimensional output embeddings.
\end{abstract}
\vspace{-0.4cm}
\section{Introduction}
A ubiquitous goal in visual representation learning is to obtain discriminative and generalizable embeddings. Such visual embeddings are learned in a deep and highly non-linear fashion. On top, a linear layer separates categories through Euclidean hyperplanes. The choice for a zero curvature Euclidean embedding space, although a \emph{de facto} standard, requires careful re-consideration as it has direct consequences for how well a task can be optimized given the latent structure that is inherently present in both the data and the category space~\cite{nickel2017poincare,liu2019hyperbolic,khrulkov2020hyperbolic}.

This work takes inspiration from recent literature advocating hyperbolic manifolds as embedding spaces for machine learning and computer vision tasks. Foundational work showed that hyperbolic manifolds are able to embed hierarchies and tree-like structures with minimal distortion~\cite{nickel2017poincare}. Follow up work has demonstrated the benefits of hyperboles for various tasks with latent hierarchical structures, from text embedding~\cite{tifrea2018poincar,zhu2020hypertext} to graph inference~\cite{chami2019hyperbolic,dai2021hyperbolic,liu2019hyperbolic}. Notably, Khrulkov \etal~\cite{khrulkov2020hyperbolic} showed that hyperbolic embeddings also have profound connections to visual data, due to latent hierarchical structures present in vision datasets. This connection has brought along early hyperbolic success in computer vision for few-shot and zero-shot learning~\cite{khrulkov2020hyperbolic,fang2021kernel,liu2020hyperbolic}, unsupervised learning~\cite{park2021unsupervised,yan2021unsupervised}, and video recognition~\cite{long2020searching,suris2021learning}.

Common amongst current hyperbolic computer vision works is that the task at hand is global, \ie an entire image or video is represented by a single vector in the hyperbolic embedding space~\cite{khrulkov2020hyperbolic,liu2020hyperbolic,long2020searching,atigh2021hyperbolic}. Here, our goal is to take hyperbolic deep learning to the pixel level. This generalization is however not trivial. The change of manifold brings different formulations for basic operations such as addition and multiplication, each with different spatial complexity. Specifically, the additional spatial complexity that comes with the M\"obius addition as part of the hyperbolic multinomial logistic regression makes it intractable to simultaneously optimize or infer all pixels of even a single image. Here, we propose an equivalent re-formulation of multinomial logistic regression in the Poincar\'e ball that bypasses the explicit computation of the M\"obius addition, allowing for simultaneous segmentation optimization on batches of images in hyperbolic space. We furthermore outline how to incorporate hierarchical knowledge amongst labels in the hyperbolic embedding space, as previously advocated in image and video recognition~\cite{liu2020hyperbolic,long2020searching}. The proposed approach is general and can be plugged on top of any segmentation architecture. \com{The code is available at \url{https://github.com/MinaGhadimiAtigh/HyperbolicImageSegmentation}.}

We perform a number of analyses to showcase the effect and new possibilities that come with Hyperbolic Image Segmentation. We present the following: \emph{(i)} Hyperbolic embeddings provide natural measures for uncertainty estimation and for semantic boundary estimation in image segmentation, see Figure~\ref{fig:fig1}. Different from Bayesian uncertainty estimation, our approach requires no additional parameters or multiple forward passes, \ie this information comes for free. \emph{(ii):} Hyperbolic embeddings with hierarchical knowledge provide better zero-label generalization than Euclidean counterparts, \ie hyperboles improve reasoning over unseen categories. \emph{(iii):} Hyperbolic embeddings are preferred for fewer embedding dimensions. Low-dimensional effectiveness is a cornerstone in hyperbolic deep learning~\cite{nickel2017poincare}. We find that these benefits extend to image segmentation, with potential for explainability and on-device segmentation~\cite{atigh2021hyperbolic}. We believe these findings bring new insights and opportunities to image segmentation.
\section{Related work}
\subsection{Image segmentation}
Widely used segmentation approaches follow the encoder-decoder paradigm, where an encoder learns lower-dimensional representations and the decoders serves to reconstruct high-resolution segmentation maps~\cite{unet15,FCN,noh2015learning,chen2017deeplab,segnet,chen18v3plus}. Early adaptations of decoders used parametrized upsampling operations through deconvolutions~\cite{FCN,noh2015learning} or multiple blocks of a bi-linear upsampling followed by more convolutional layers~\cite{segnet}. More recent works seek to reinforce the upsampling with context information by merging feature maps at various scales, \ie feature pyramids~\cite{zhao2017pyramid}, or by combining the decoding with global context features through fully connected layers \cite{zhang2018context}. For example, the widely adapted Deeplab architecture~\cite{chen2017deeplab} uses atrous convolutions with various levels of dilation within the decoder to effectively obtain context information at various scales.
Other recent approaches focus on improving the utilization of multi-scale information, \eg using multi-scale attention~\cite{tao2020hierarchical}, squeeze-and-attention~\cite{Zhong_2020_CVPR}, and Transformers~\cite{YuanCW20}.
Commonly in semantic image segmentation\com{,} the final classification is performed through multinomial logistic regression in Euclidean space. As a promising alternative, we advocate for using the hyperbolic space to perform pixel-level classification on top of any existing architecture.
\subsection{Hyperbolic deep learning}
The hyperbolic space has gained traction in deep learning literature for representing tree-like structures and taxonomies~\cite{ganea2018hyperbolic,law2019lorentzian,nickel2017poincare,nickel2018learning,sarkar2011low,sala2018representation,yu2019numerically}, text~\cite{aly2019every,tifrea2018poincar,zhu2020hypertext}, and graphs~\cite{bachmann2020constant,chami2019hyperbolic,dai2021hyperbolic,liu2019hyperbolic,lou2020differentiating,zhang2021lorentzian}. Hyperbolic alternatives have been proposed for various network layers, from intermediate layers~\cite{ganea2018hyperbolic2,shimizu2021hyperbolic} to classification layers~\cite{cho2019large,ganea2018hyperbolic2,atigh2021hyperbolic,shimizu2021hyperbolic}. Recently, hyperboles have also been applied in computer vision for hierarchical action search \cite{long2020searching}, few-shot learning \cite{khrulkov2020hyperbolic}, hierarchical image classification \cite{dhall2020hierarchical}, and zero-shot image recognition~\cite{liu2020hyperbolic}. In this work, we build upon these foundations and make the step towards semantic image segmentation, which requires a reformulation of the hyperbolic multinomial logistic regression to become tractable.

Previous works have shown the potential of a hierarchical view on image segmentation.
For instance, \cite{Zhao_2017_ICCV} incorporate an open vocabulary perspective based on WordNet \cite{Miller95wordnet:a} hypernym/hyponym relations. By learning a joint-embedding of image features and word concepts, combined with a dedicated scoring function to enforce the asymmetric relation between hypernyms and hyponyms, their model is able to predict hierarchical concepts. This approach is akin to that of \cite{liang2018b} who use hierarchy-level specific convolutional blocks. These blocks, individually tasked with discriminating only between child classes, are dynamically activated such that only a subset of the entire graph is activated at any given time depending on which concepts are present in the image. This is trained with a loss function consisting of a sum of binary cross-entropy losses at each of the child-concept prediction maps.
Here, we seek to incorporate hierarchical information on the hyperbolic manifolds, which can be applied on top of any segmentation architecture without needing to change the architecture itself.

Recent work by \cite{weng2021unsupervised} investigated the use of the hyperbolic space for instance segmentation in images, but only do so after the fact, \ie on top of predicted instance segmentations. In contrast, our approach enables tractable hyperbolic classification as part of the pixel-level segmentation itself.
\section{Image segmentation on the hyperbole}

\subsection{Background: The Poincar\'e ball model}
Hyperbolic geometry encompases several conformal models~\cite{cannon1997hyperbolic}. Based on its widespread use in deep learning and computer vision, we operate on the Poincar\'{e} ball. The Poincar\'{e} ball is defined as $(\mathbb{D}^n_c, g^{\mathbb{D}_c})$, with manifold $\mathbb{D}^n_c = \{x \in \mathbb{R}^n : c||x|| < 1\}$ and Riemannian metric:
\begin{equation}
g^{\mathbb{D}_c}_x = (\lambda_x^c)^2 g^{E} = \frac{2}{1 - c||x||^2} \mathbb{I}^n,
\label{eq:poincaremetric}
\end{equation}
where \com{$g^{E}=\mathbb{I}^n$} denotes the Euclidean metric tensor and $c$ is a hyperparameter governing the curvature and radius of the ball. Segmentation networks operate in Euclidean space and to be able to operate on the Poincar\'{e} ball, a mapping from the Euclidean tangent space to the hyperbolic space is required. The projection of a Euclidean vector $x$ onto the Poincar\'{e} ball is given by the exponential map with anchor $v$:
\begin{equation}
\exp^c_{v}(x) = v \oplus_c \bigg(\tanh \bigg(\sqrt{c} \frac{\lambda_v^c||x||}{2} \bigg)\frac{x}{\sqrt{c}||x||} \bigg),
\end{equation}
with $\oplus_c$ the M\"{o}bius addition:
\begin{equation}
v \oplus_c w = \frac{(1 + 2c \langle v, w \rangle + c||w||^2)v + (1 - c||v||^2)w}{1 + 2c \langle v, w \rangle + c^2||v||^2 ||w||^2}.
\end{equation}
In practice, $\mathbf{v}$ is commonly set to the origin, simplifying the exponential map to \begin{equation}
\exp_{0}(x) = \tanh (\sqrt{c}||x||) (x/(\sqrt{c}||x||)).
\end{equation}

\subsection{Tractable pixel-level hyperbolic classification}
For the problem of image segmentation, we are given an input image $X \in \mathbb{R}^{w \times h \times 3}$, with $w$ and $h$ the width and height of the image respectively. For each pixel $x \in X$, we need to assign a label $y \in Y$, where $Y$ denotes a set of $C$ class labels. Let $f(X): \mathbb{R}^{w \times h \times 3} \mapsto \mathbb{R}^{w \times h \times \com{n}}$ denote an arbitrary function that transforms each pixel to a \com{$n$}-dimensional representation, \eg an image-to-image network. Common amongst current approaches is to feed all pixels in parallel to a linear layer followed by a softmax, resulting in a $C$-dimensional probability distribution over all $C$ classes per pixel, optimized with cross-entropy.

\begin{figure}[t]
\centering
    \includegraphics[width=0.32\textwidth]{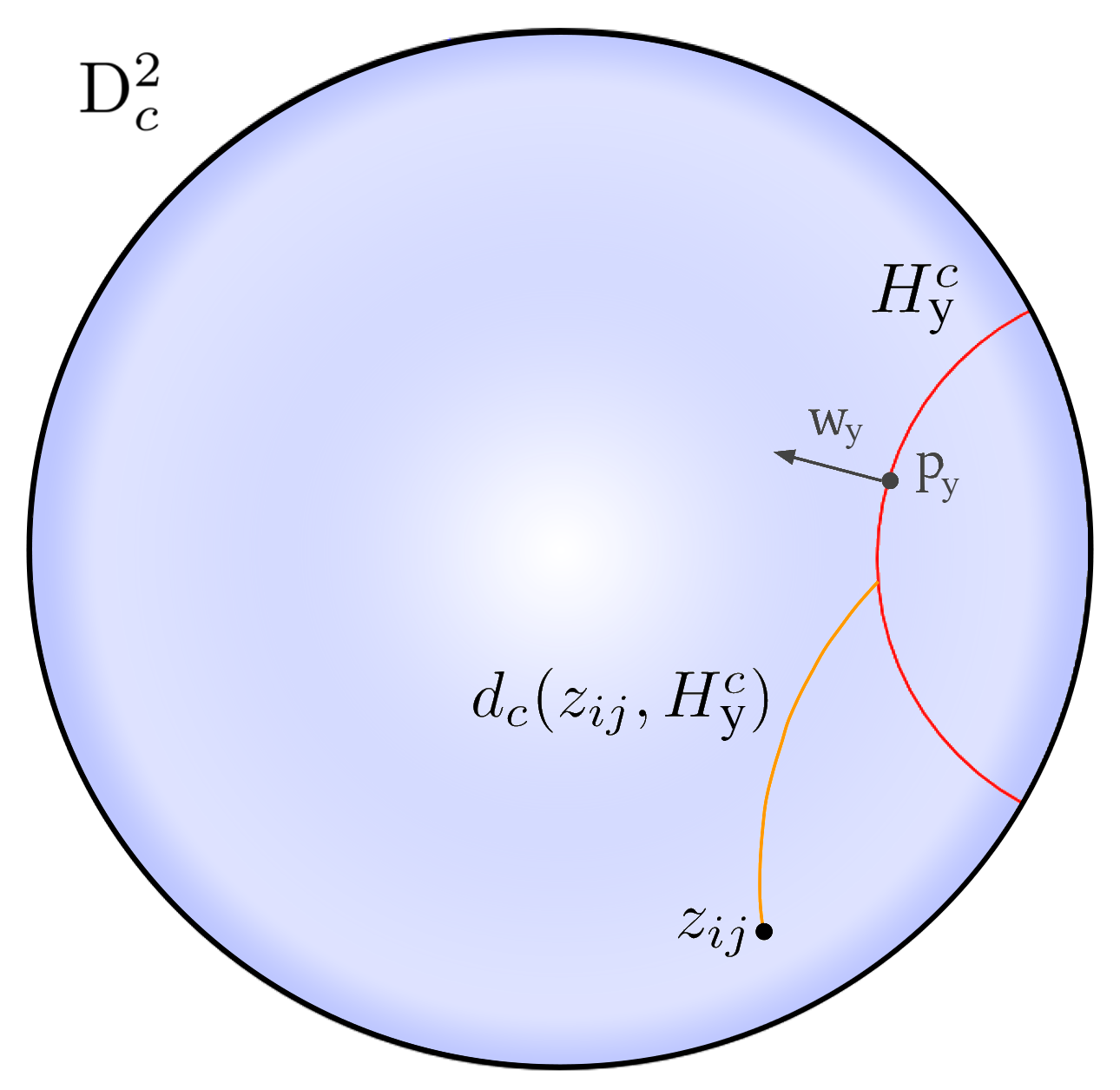}
   \caption{Visualization of the hyperbolic gyroplane $(p_y, w_y)$ and distance to output $z_{ij}$ on a two-dimensional manifold. In the context of this work, $z_{ij}$ denotes the output representation at pixel location $(i,j)$ and $H^c_y$ denotes the hyperplane for class $y$.} 
\label{fig:method}
\vspace{-0.4cm}
\end{figure}

This paper advocates the use of the hyperbolic space to perform the per-pixel classification for image segmentation. We start from the geometric interpretation of the hyperbolic multinomial logistic regression given by Ganea \etal~\cite{ganea2018hyperbolic}, which defines the gyroplane, \ie the hyperplane in the Poincar\'{e} ball, as:
\begin{equation}
H^c = \{z_{ij} \in \mathbb{D}^n_c, \langle -p \oplus_c z_{ij}, w \rangle = 0 \},
\end{equation}
where $z_{ij} = \exp_0(f(X)_{ij})$ denotes the exponential map of the network output at pixel location $(i,j)$ and with $p \in \mathbb{D}^n_c$ the offset and $w \in \mathcal{T}_p\mathbb{D}^n_c$ the orientation of the gyroplane. The hyperbolic distance of $z_{ij}$ to the gyroplane of class $y$ is given as:
\begin{equation}
d_c(z_{ij}, H^c_y) = \frac{1}{\sqrt{c}}\sinh^{-1} \bigg( \frac{2 \sqrt{c} \langle -p_y \oplus_c z_{ij}, w_y \rangle}{(1\!-\!c ||-p_y\! \oplus_c\! z_{ij}||^2)||w_y||} \bigg).
\end{equation}
Figure~\ref{fig:method} illustrates a gyroplane on the hyperbole defined by its offset and orientation, along with the geodesic from pixel output $z_{ij}$ to the gyroplane. Based on this distance, the logit of class $y$ for pixel output $z_{ij}$ using the metric of Equation~\ref{eq:poincaremetric} is given as:
\begin{equation}
\zeta_y(z_{ij}) = \frac{\lambda^c_{p_y}||w_y||}{\sqrt{c}} \sinh^{-1} \bigg( \frac{2 \sqrt{c} \langle -p_y \oplus_c z_{ij}, w_y \rangle}{(1\!-\!c ||-p_y\! \oplus_c\! z_{ij}||^2)||w_y||} \bigg).
\label{eq:logit}
\end{equation}
Consequently, the likelihood is given as:
\begin{equation}
p(\hat{y} = y|z_{ij}) \propto \exp(\zeta_y(z_{ij})),
\end{equation}
which can be optimized with the cross-entropy loss and gradient descent.

The geometric interpretation of Ganea \etal~\cite{ganea2018hyperbolic} provides a framework for classifying output vectors in hyperbolic space. In contrast to standard classification, image segmentation requires per-pixel classification in parallel. This setup is however intractable for the current implementation of hyperbolic multinomial logistic regression. The bottleneck is formed by the explicit computation of the M\"{o}bius addition. In a standard example segmentation setting ($W=H=513$, $K=100$ classes, $n=256$, and batch size $5$), this would induce a memory footprint of roughly 132 GB in 32-bit float precision, compared to roughly 0.5 GB in Euclidean space. Here, we propose an equivalent computation of the margin likelihood by factoring out the explicit computation of the M\"{o}bius addition\com{, resulting in a memory footprint of 1.1 GB}. The key to our approach is the observation that we do not need the actual result of the addition, only its inner product in the numerator of Equation~\ref{eq:logit} $\langle -p_y \oplus_c z_{ij}, w_y \rangle$
and its squared norm in the denominator $||-p_y \oplus_c z_{ij}||^2$.

To that end, we first rewrite the M\"{o}bius addition as:
\begin{equation}
\begin{split}
\hat{p}_y \oplus_c z_{ij} &= \alpha \hat{p}_y + \beta z_{ij},\\
 \alpha &= \frac{1 + 2c \langle \hat{p}_y, z_{ij} \rangle + c ||z_{ij}||^2}{1 + 2c \langle \hat{p}_y, z_{ij} \rangle + c^2 ||\hat{p}_y||^2 ||z_{ij}||^2},\\
 \beta &= \frac{1 - c ||\hat{p}_y||^2}{1 + 2c \langle \hat{p}_y, z_{ij} \rangle + c^2 ||\hat{p}_y||^2 ||z_{ij}||^2}.
\end{split}
\end{equation}
with $\hat{p}_y = -p_y$ for clarity.
The formulation above allows us to precompute $\alpha$ and $\beta$ for reuse. Then, we rewrite the inner product with $w_y$ as:
\begin{equation}
\begin{split}
\langle \hat{p}_y \oplus_c z_{ij}, w_y \rangle & = \langle \alpha \hat{p}_y + \beta z_{ij}, w_y \rangle,\\
& = \alpha \langle \hat{p}_y, w \rangle + \beta \langle z_{ij}, w \rangle.
\end{split}
\end{equation}
Where an explicit computation of the M\"{o}bius addition requires evaluating a tensor in $\mathbb{R}^{W \times H \times C \times n}$ for a single image, this is reduced to adding two tensors in $\mathbb{R}^{W \times H \times C}$. The squared norm of the M\"{o}bius addition can be efficiently computed as follows:
\begin{equation}
\begin{split}
||\hat{p}_y & \oplus_c z_{ij}||^2 = \sum_{m=1}^{n} (\alpha \hat{p}_y^{m} + \beta z_{ij}^{m})^2,\\
&= \sum_{m=1}^{n} (\alpha \hat{p}_y^{m})^2 + \alpha \hat{p}_y^{m} \beta z_{ij}^{m} + (\beta z_{ij}^{m})^2,\\
&= \alpha^2 \sum_{m=1}^{n} (\hat{p}_y^{m})^2 + 2 \alpha \beta \sum_{m=1}^{n} \hat{p}_y^{m} z_{ij}^{m} + \beta^2 \sum_{m=1}^{n} (z_{ij}^{m})^2,\\
&= \alpha^2||\hat{p}_y||^2 +  2 \alpha \beta \langle \hat{p}_y, z_{ij} \rangle + \beta^2 ||z_{ij}||^2,
\end{split}
\end{equation}
which is a summation of three tensors in $\mathbb{R}^{W \times H \times C}$. Moreover, all terms have already been computed when precomputing $\alpha$ and $\beta$. By the reformulation of the inner product and squared norm when computing the class logits, we make hyperbolic classification feasible at the pixel level.

\begin{figure}[t]
\centering
{
    \includegraphics[trim=25 60 25 80,clip=true,width=0.45\textwidth]{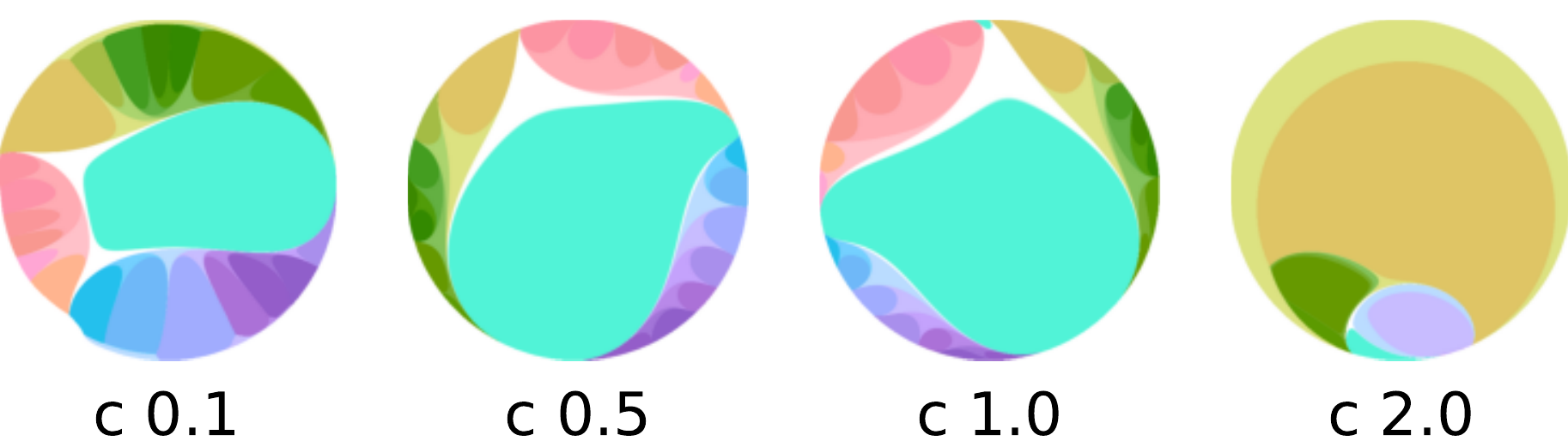}
}
  \caption{\textbf{Visualizing class embeddings} in hyperbolic space for the 20 classes of Pascal VOC. The colors outline the hierarchical structure of the classes. The higher the curvature, the more the gyroplanes are positioned towards the edge of the Poincar\'e disk. In the analyses, we investigate the quantitative effect of hyperbolic curvature for segmentation performance.} 
\label{fig:discs}
\end{figure}

\subsection{Hierarchical hyperbolic class embedding}
It has been repeatedly shown that the hyperbolic space is able to embed hierarchical structures with minimal distortion~\cite{peng2021hyperbolic,sala2018representation,sarkar2011low}. To that end, we investigate the potential of incorporating hierarchical relations between classes for image segmentation on hyperbolic manifolds. Let $Y$ denote the set of all classes, which form the leaf nodes of hierarchy $\mathcal{N}$. For class $y \in Y$, let $\mathcal{A}_y$ denote the ancestors of $y$. The probability of class $y$ for output $z_{ij}$ is then given by a hierarchical softmax:
\begin{equation}
\begin{split}
p(\hat{y} = y | z_{ij}) & =
\prod_{h \in \mathcal{H}_y} p(h | \mathcal{A}_h, z_{ij})\\ & = \prod_{h \in \mathcal{H}_y} \frac{\exp(\zeta_h(z_{ij}))}{\sum_{s \in S_h} \exp(\zeta_{s}(z_{ij}))},
\end{split}
\label{eq:jointprob}
\end{equation}
with $\mathcal{H}_y = \{y\} \cup \mathcal{A}_y$ and with $S_h$ the siblings of $h$. The above formulation calculates the joint probability from root to leaf node, where the probability at each node is given as the softmax normalized by the siblings in the same subtree. Given this probability function, training can be performed with cross-entropy and the most likely class is selected during inference based on Equation~\ref{eq:jointprob}. In Figure~\ref{fig:discs}, we visualize how incorporating such knowledge results in a hierarchically consistent embedding of class gyroplanes.
\section{Analyses}

\begin{figure*}[t]
\centering
\begin{subfigure}{0.49\textwidth}
\includegraphics[trim=10 15 15 13,clip=true, width=\textwidth]{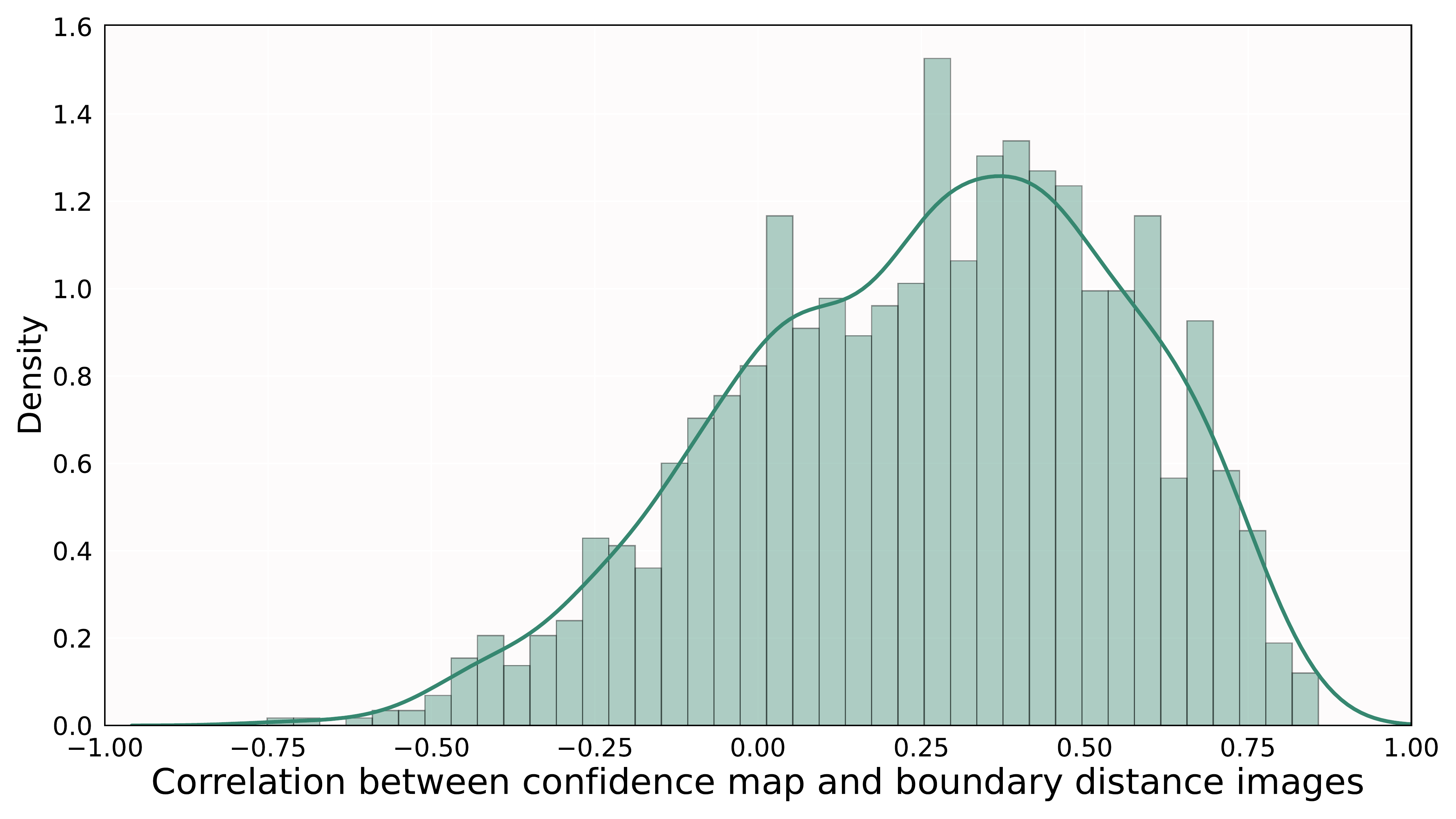}
\caption{Hyperbolic uncertainty correlates with boundary distance.}
\label{fig:border}
\end{subfigure}
\begin{subfigure}{0.49\textwidth}
\includegraphics[trim=10 15 10 4,clip=true, width=\textwidth]{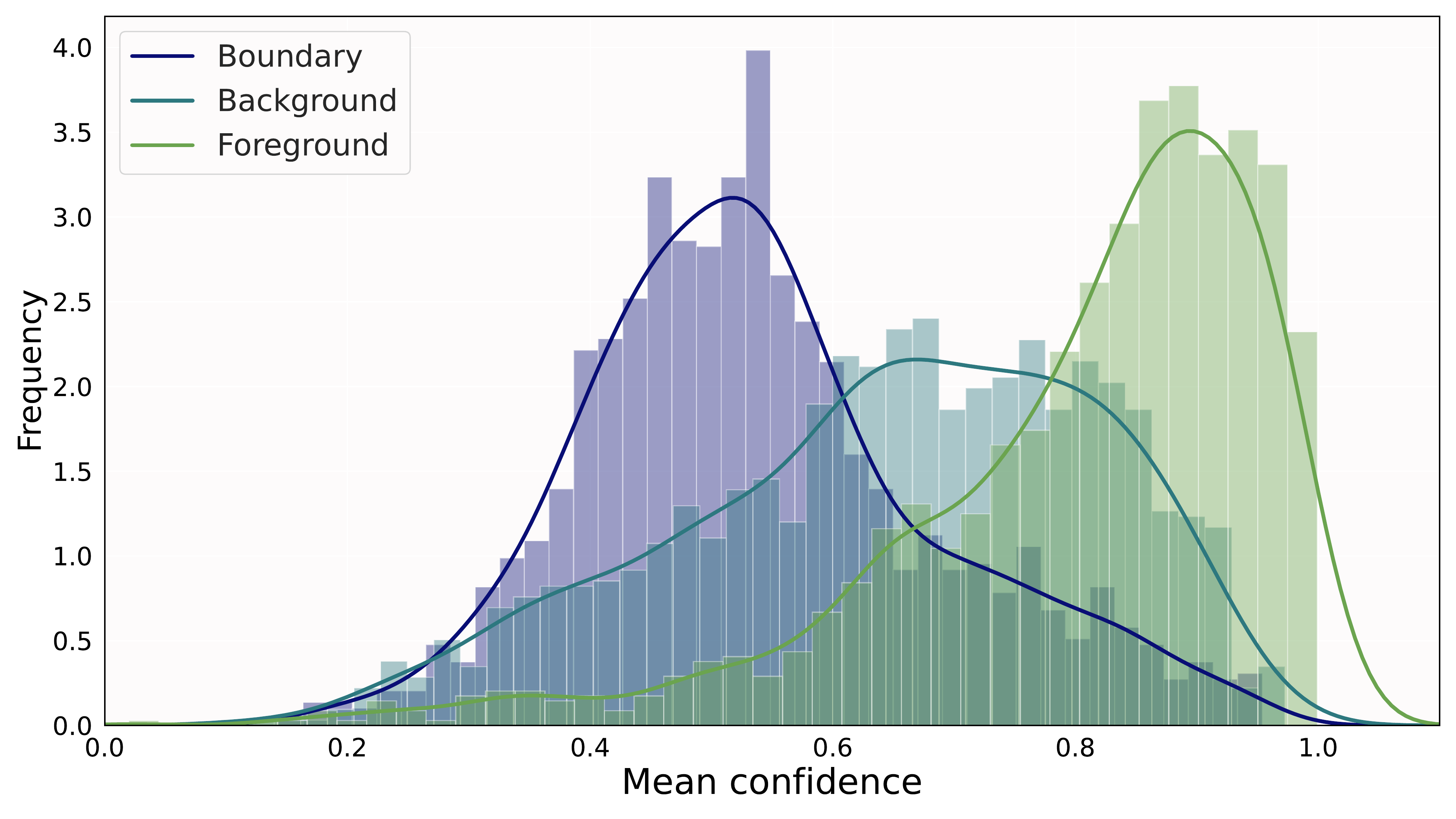}
\caption{Hyperbolic uncertainty is higher for boundaries than object interiors.}
\label{fig:border2}
\end{subfigure}
\vspace{-0.4cm}
\caption{\textbf{Is hyperbolic uncertainty semantically meaningful?} We perform two quantitative experiments on Pascal VOC with 2 embedding dimensions to uncover whether hyperbolic uncertainty provides meaningful insights. Left: we find that the per-pixel hyperbolic uncertainty (here shown as its inverse, namely confidence) strongly correlates with semantic boundaries in the segmentation. Right: hyperbolic confidence is highest for foreground pixels denoting object interiors, followed by background pixels and finally semantic boundaries.}
\end{figure*}

\subsection{Setup}
\noindent
\textbf{Datasets.}
We evaluate Hyperbolic Image Segmentation on three datasets, COCO-Stuff-10K~\cite{caesar2018coco}, Pascal VOC~\cite{everingham2015pascal}, and ADE20K~\cite{zhou2017scene}.
\textbf{COCO-Stuff-10K} contains 10,000 images from 171 classes consisting of 80 countable \emph{thing} classes such as \emph{umbrella} or \emph{car}, and 91 uncountable \emph{stuff} classes such as \emph{sky} or \emph{water}. The dataset is split into 9,000 images in the training set and 1,000 images in the test set.
\textbf{Pascal VOC} contains 12,031 images from 21 classes consisting of 20 object classes like \emph{person} and \emph{sheep} and a \emph{background} class. The dataset is split into 10,582 images in the train set and 1,449 images in the test set.
\textbf{ADE20K} contains 22,210 images from 150 classes, such as \emph{car} and \emph{water}. The dataset is split into 20,210 in the train set and 2000  images in the test set. For all datasets, we have made the full hierarchies and they are shown in the supplementary materials.

\textbf{Implementation details.}
For all experiments, we use DeeplabV3+ with a ResNet101 backbone~\cite{chen18v3plus}. We initialize the learning rate to be 0.001, 0.001, and 0.01 for COCO-stuff-10k, ADE20K, and Pascal VOC. We train the model for 70, 140, and 40 epochs for COCO-stuff-10K, ADE20K, and Pascal VOC with a batch size of 5.
To optimize Euclidean parameters, we use SGD with a momentum of 0.9 and polynomial learning rate decay with a power of 0.9 akin to~\cite{chen18v3plus}. To optimize Hyperbolic parameters, we use RSGD, similar to~\cite{ganea2018hyperbolic}.

\textbf{Evaluation metrics.}
We perform the evaluation on both standard and hierarchical metrics.
For the standard metrics, we use pixel accuracy (PA), class accuracy (CA), and mean Intersection Over Union (mIOU). Pixel accuracy denotes the percentage of pixels in the image with the correct label. Class accuracy first calculates the accuracy per class and then averages over all classes. IOU denotes the spatial overlap of ground truth and predicted segmentation. mIOU denotes the mean IOU over all classes.
To evaluate hierarchical consistency and robustness, we also report sibling and cousin variants of each metric, following~\cite{long2020searching}. In the sibling variant of the metrics, a prediction is also counted as correct if it shares a parent with the target class. In the cousin variants, the predicted labels need to share a grandparent with the target class to count as correct.

\subsection{Uncertainty and boundary information for free}
The ability to interpret predictions is vital in many segmentation scenarios, from medical imaging to autonomous driving, to invoke trust and enable decision making with a human in the loop~\cite{akata2020hybrid}. For the first analysis, we investigate the role of hyperbolic embeddings for interpretation in segmentation. Specifically, we show how the distance to the origin of each pixel in the hyperbolic embedding space provides a natural measure of uncertainty prediction. We draw comparisons to Bayesian uncertainty and investigate whether hyperbolic uncertainty is semantically meaningful.


\begin{figure}[t]
\centering
    \includegraphics[trim=345 7 0 0,clip=true, width=\linewidth]{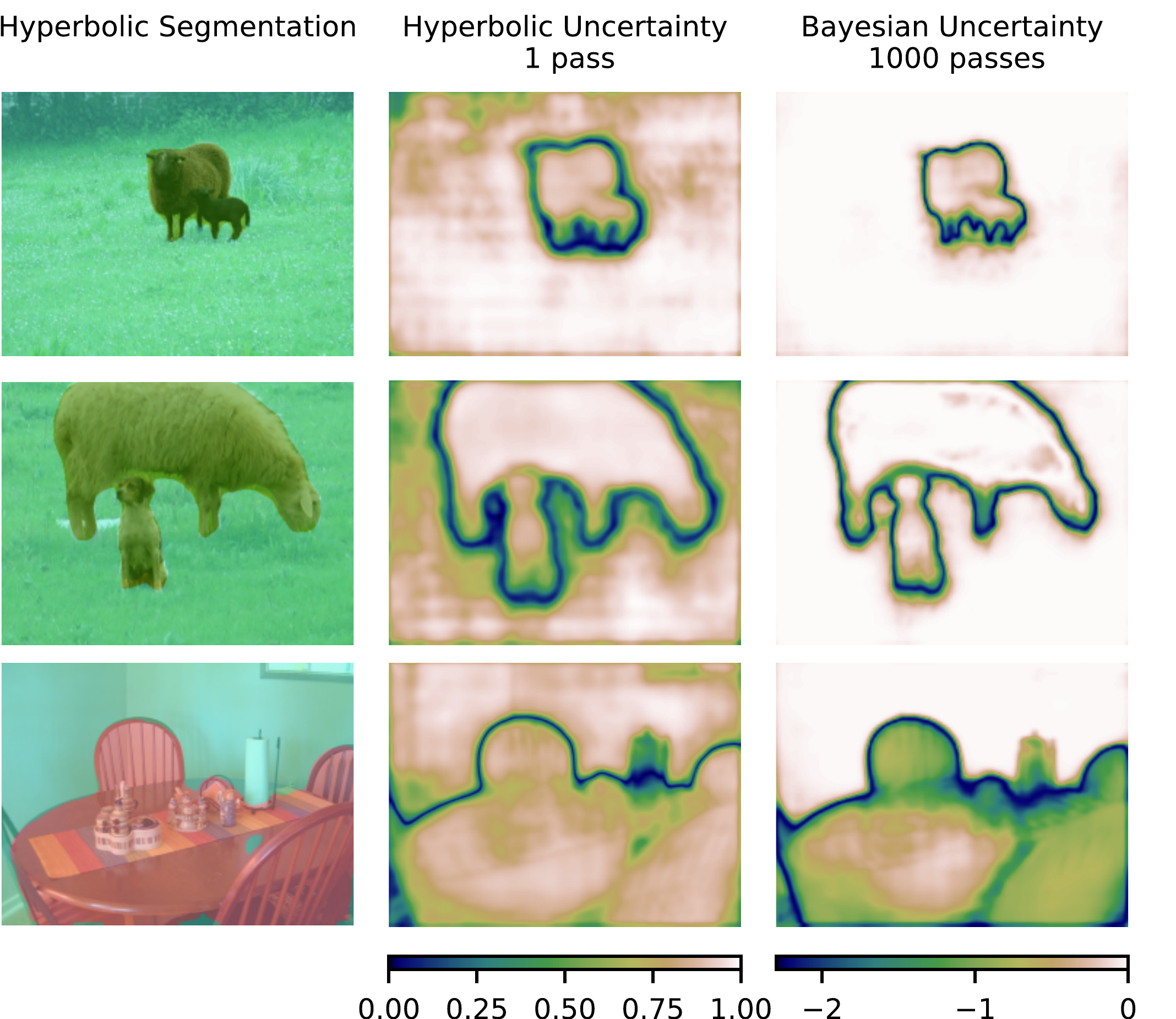}
  \caption{\textbf{Hyperbolic vs Euclidean uncertainty} for examples from Pascal VOC. Both measures of uncertainty are highly aligned and focus on semantic boundaries. However, the Bayesian uncertainty for Euclidean embeddings require 1,000 passes, whereas we obtain uncertainty for free with hyperbolic embeddings.}
\label{fig:bayesian}
\vspace{-0.5cm}
\end{figure}

\textbf{Hyperbolic vs Bayesian uncertainty.}
To obtain per-pixel uncertainty in Hyperbolic Image Segmentation, we simply measure the $\ell_2$ norm to the origin in the Poincar\'e ball, regardless of their positioning to the class-specific gyroplanes. In conventional segmentation architectures, such uncertainty measures are more commonly obtained through Bayesian optimization, either by making the network Bayesian from the start~\cite{tran2019bayesianlayers} or through Monte-Carlo dropout during inference~\cite{mukhoti2018evaluating}.

In Figure~\ref{fig:bayesian}, we show the uncertainty maps for examples from Pascal VOC for hyperbolic uncertainty with 2 embedding dimensions and curvature 0.1. We draw a qualitative comparison to its Bayesian counterpart in Euclidean space by way of dropout during inference~\cite{gal2016baydropout}. Both variants employ the same backbone. 
To create the Bayesian uncertainty map, we add Mont-Carlo dropout after Resnet blocks with a drop ratio of 0.5 and pass each image 1,000 times through the network, similar to~\cite{mukhoti2018evaluating}.
Figure~\ref{fig:bayesian} shows three hyperbolic and Bayesian uncertainty example maps. Both uncertainty maps are highly interpretable, focusing on semantic boundaries and occluded areas of the image. A key difference however is the amount of network passes required to obtain the maps: 
\com{while} Bayesian uncertainty requires many passes due to the MC dropout, we obtain the uncertainty maps for free, resulting in a 1,000-fold inference speed-up.

\textbf{Is hyperbolic uncertainty semantically meaningful?} The qualitative results suggest that the hyperbolic uncertainty measure is semantically meaningful, as it relates to the semantic boundaries between objects. To test this hypothesis, we have outlined a quantitative experiment: for each pixel in the ground truth segmentation map, we compute the Euclidean distance to the nearest pixel with another class label. Intuitively, this distance correlates with prediction confidence; the closer to the boundary, the smaller the hyperbolic norm. We perform a correlation analysis between confidence and boundary distance for all pixels in an image. We then aggregate the correlations over all images.

In Figure~\ref{fig:border}, we show a histogram of the correlations over all images in Pascal VOC with the same embedding dimensionality and curvature as above. The histogram shows that the confidence (inverse of uncertainty) from our hyperbolic approach clearly correlates with the distance to the nearest boundary. This result highlights that hyperbolic uncertainty provides a direct clue about which regions in the image contain boundaries between images, which can\com{,} in turn\com{,} be used to determine whether to ignore such regions or to pinpoint where to optimize further as boundary areas commonly contain many errors~\cite{samson2019bet}. We provide the same experiment for 256 embedding dimensions in the supplementary materials, which follows the same distribution.

To further highlight the relation between hyperbolic uncertainty and semantic boundaries, we have performed a second quantitative experiment, where we classify each pixel into one of three classes: boundary pixel if it is within 10 distance\com{s} from the nearest other class, background pixel, or foreground pixel (\ie one of the other objects). In Figure~\ref{fig:border2}, we plot the mean confidence per pixel on Pascal VOC over all three classes, showing that hyperbolic confidence is highest for foreground pixels and lowest for boundary pixels, with background pixels in between. All information about boundaries and pixel classes come\com{s} for free with hyperboles as the embedding space in segmentation.

\begin{table}[t]
\resizebox{1\linewidth}{!}{
\begin{tabular}{ccccc}
\toprule
\multicolumn{5}{c}{\textbf{COCO-Stuff-10k}}\\
Manifold & Hierarchical & Class Acc & Pixel Acc & mIOU\\
\midrule
$\mathbb{R}$ & & 0.44 & 0.33 & 0.23\\
$\mathbb{R}$ & $\checkmark$ & 3.29 & 48.65 & 18.53\\
\rowcolor{Gray}
$\mathbb{D}$ & $\checkmark$ & \textbf{3.46} & \textbf{51.70} & \textbf{21.15}\\
\bottomrule
\end{tabular}
}
\vspace{0.5cm}\\
\resizebox{1\linewidth}{!}{
\begin{tabular}{ccccc}
\toprule
\multicolumn{5}{c}{\textbf{Pascal VOC}}\\
Manifold & Hierarchical & Class Acc & Pixel Acc & mIOU\\
\midrule
$\mathbb{R}$ & & 4.88 & 10.84 & 2.59\\
$\mathbb{R}$ & $\checkmark$ & 7.80 & 31.04 & 16.15\\
\rowcolor{Gray}
$\mathbb{D}$ & $\checkmark$ & \textbf{12.15} & \textbf{47.92} & \textbf{34.87}\\
\bottomrule
\end{tabular}
}
\caption{\textbf{Zero-label generalization} on Coco-Stuff-10k and Pascal VOC. On both datasets, combining hierarchical knowledge with hyperbolic embeddings provides a more suitable foundation for generalizing to unseen classes than its Euclidean counterpart.}
\label{tab:zerolabel}
\vspace{-0.5cm}
\end{table}

\subsection{Zero-label generalization}
In the second analysis, we demonstrate the potential of hyperbolic embeddings to generalize to unseen classes for image segmentation. We perform zero-label experiments on COCO-Stuff-10k and Pascal VOC and follow the zero-label semantic segmentation setup from Xian \etal~\cite{xian2019semantic}. For COCO-Stuff-10k we use a set of 15 unseen classes for inference, corresponding to all classes in the dataset that do not occur in the 2014 ImageNet Large Scale Visual Recognition Challenge~\cite{russakovsky2015imagenet}, on which the backbone was pre-trained. This assures that the model has never seen any of the classes during training. For Pascal VOC, we follow the 15/5 seen/unseen split of~\cite{xian2019semantic}. We draw a comparison to two baselines: the standard DeepLabV3+, which operates in Euclidean space and does not employ hierarchical relations, and a variant of DeepLabV3+ that employs a Euclidean hierarchical softmax.

\begin{figure*}[t]
\centering
    \includegraphics[trim=25 31 23 26,clip=true, width=0.9\textwidth]{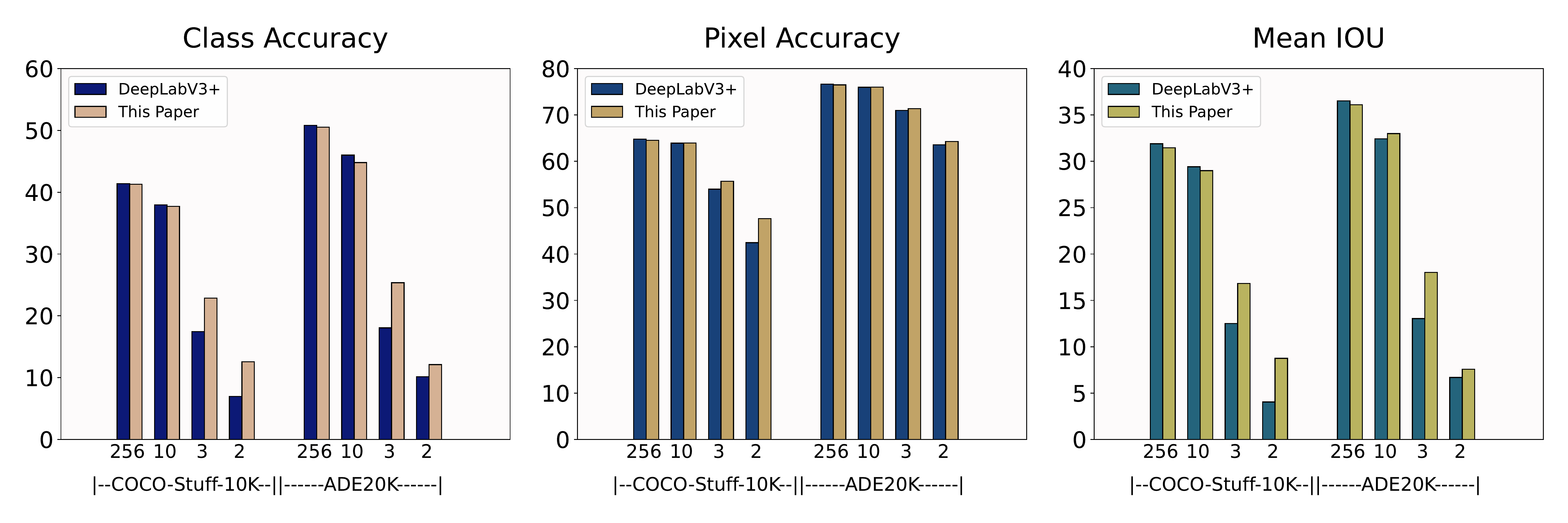}
   \caption{\textbf{Low-dimensional effectiveness of hyperbolic embeddings} for image segmentation on COCO-Stuff-10k and ADE20k. Across all three metrics, our approach obtains competitive performance in high-dimensional embedding spaces to the Euclidean counterpart. When restricting the embedding space to a few dimensions, hyperbolic embeddings are preferred for segmentation.}
\label{fig:dimension}
\end{figure*}

\begin{figure*}[t]
\centering
    \includegraphics[trim=10 55 10 55,clip=true, width=0.975\linewidth]{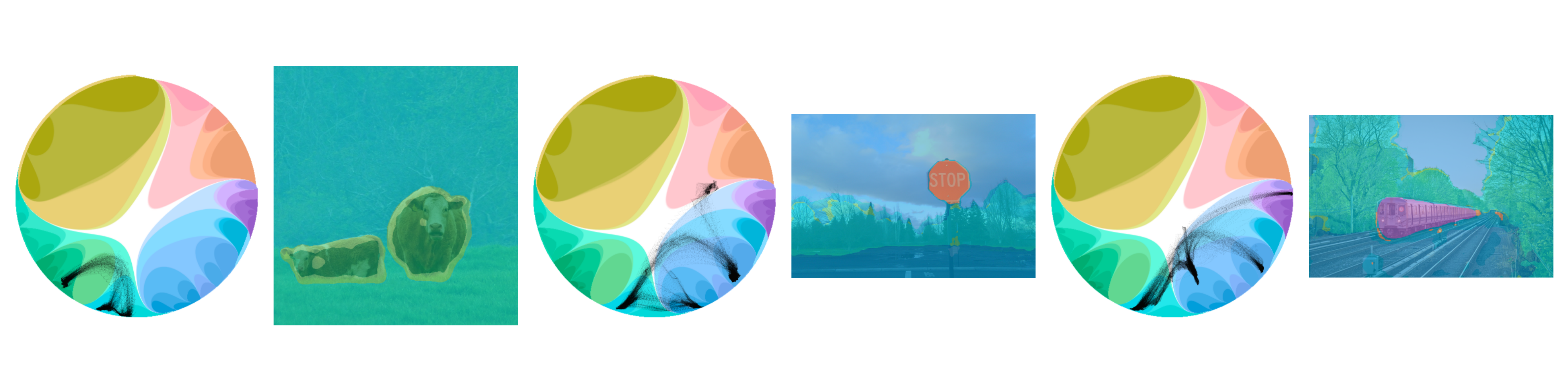}
  \caption{\textbf{Qualitative examples of Hyperbolic Image Segmentation} with two embedding dimensions on COCO-Stuff-10k. For each example, we show the projection of all pixels in the hyperbolic embedding (left) and the segmentation result (right). From left to right: the lime color denotes \emph{cow} (partial failure case), the red color denotes \emph{stop sign}, and the purple color denotes \emph{train}.
  }
\label{fig:samples}
\vspace{-0.4cm}
\end{figure*}

More formally, given a set of unseen classes $C_U$ and a set of seen classes $C_S$, we remove all $k \in C_U$ from the dataset by replacing them with an ignore label. This effectively means that these pixels are not used during optimization and the model is therefore not optimized on these classes. As such, in images containing concepts from $C_U$, the pixels containing the concepts from $C_S$ are still used in training. Different from the more widely known zero-shot image classification task, images containing unseen concepts are not removed from the training set. Removing these images would result in a significantly reduced training set, which is impractical for the purposes of the evaluation. After training on $C_S$, we perform inference by choosing only between unseen concepts for each pixel. We note that we do not adapt our approach to the zero-label setting, we employ the same network and loss as for supervised segmentation, the only difference lies in the used classes for training and inference.

The results on COCO-Stuff-10k and Pascal VOC are shown in Table~\ref{tab:zerolabel} for 256 output dimensions and respective curvatures 1 and 2. In the supplementary materials, we also show the results using the sibling and cousin variants of the three metrics. For both datasets, we first observe that using a standard Euclidean architecture without hierarchical knowledge results in near-random zero-label performance. When using hierarchical knowledge and Euclidean embeddings, it becomes possible to recognize unseen classes. To generalize towards unseen classes\com{,} however, it is best to combine class hierarchies with hyperbolic embeddings. On COCO-Stuff-10k, the mIOU increases from 18.53 to 20.76. On Pascal VOC, the difference is even bigger; from 16.15 to 34.87. This experiment shows the strong affinity between hierarchical knowledge and hyperbolic embeddings for image segmentation and the potential for generalizing to unseen classes. We conclude that the hyperbolic space provides a more suitable foundation for generalizing to unseen classes in the context of segmentation. Qualitative zero-label results are provided in the supplementary materials.

\subsection{Low-dimensional embedding effectiveness}
In the third analysis, we demonstrate the effectiveness of hyperbolic embeddings in a low-dimensional setting. Hyperboles have shown to be beneficial with few embedding dimensions on various data types. In Figure~\ref{fig:dimension}, we compare the default Euclidean embeddings to hyperbolic embeddings for DeepLabV3+ on COCO-Stuff-10k and ADE20K, with a dimensionality ranging from $256$ to $2$. The standard setting \com{of classical segmentation} for DeepLabV3+ is to operate on a dimensionality of $256$. Low dimensional embeddings are however preferred for explainability and on-device segmentation~\cite{atigh2021hyperbolic}, due to their reduced complexity and smaller memory footprint.

Our results show a consistent pattern across both datasets and the 
metrics, where hyperbolic embeddings obtain comparable performances for high ($256$) or medium ($10$) dimension settings. In low-dimensional settings ($2$ and $3$), our approach outperforms DeepLabV3+. As expected, the performance of both models drops when using lower dimensional embeddings, but as is especially apparent on the COCO-Stuff-10k dataset, the Euclidean default is affected most. By using a structured embedding space we are able to obtain better performance in low-dimensions, for as low as $2$ dimensions. When using $3$ dimensions, hyperbolic embeddings improve the mIOU by $4.32$ percent point on COCO-Stuff-10k and by $4.99$ on ADE20k. The benefits of this low dimensional embedding for explainability are demonstrated with the hyperbolic disk visualisations in this paper, which are based on models trained in $2$ dimensions. We conclude that the low-dimensional effectiveness of hyperbolic embeddings extend\com{s} to the task of image segmentation. In Figure~\ref{fig:samples} we provide qualitative examples in 2-dimensional hyperbolic embedding spaces. \com{Further explanation on the colors is provided in the supplementary materials.}

\subsection{Further ablations}
To complete the analyses, we ablate two design choices in our approach, namely the hyperbolic curvature and the use of hierarchical relations in the hyperbolic embedding space. Both ablations are performed on COCO-Stuff-10k.

\begin{figure}[t]
\centering
    \includegraphics[trim=25 30 24 24,clip=true, width=\linewidth]{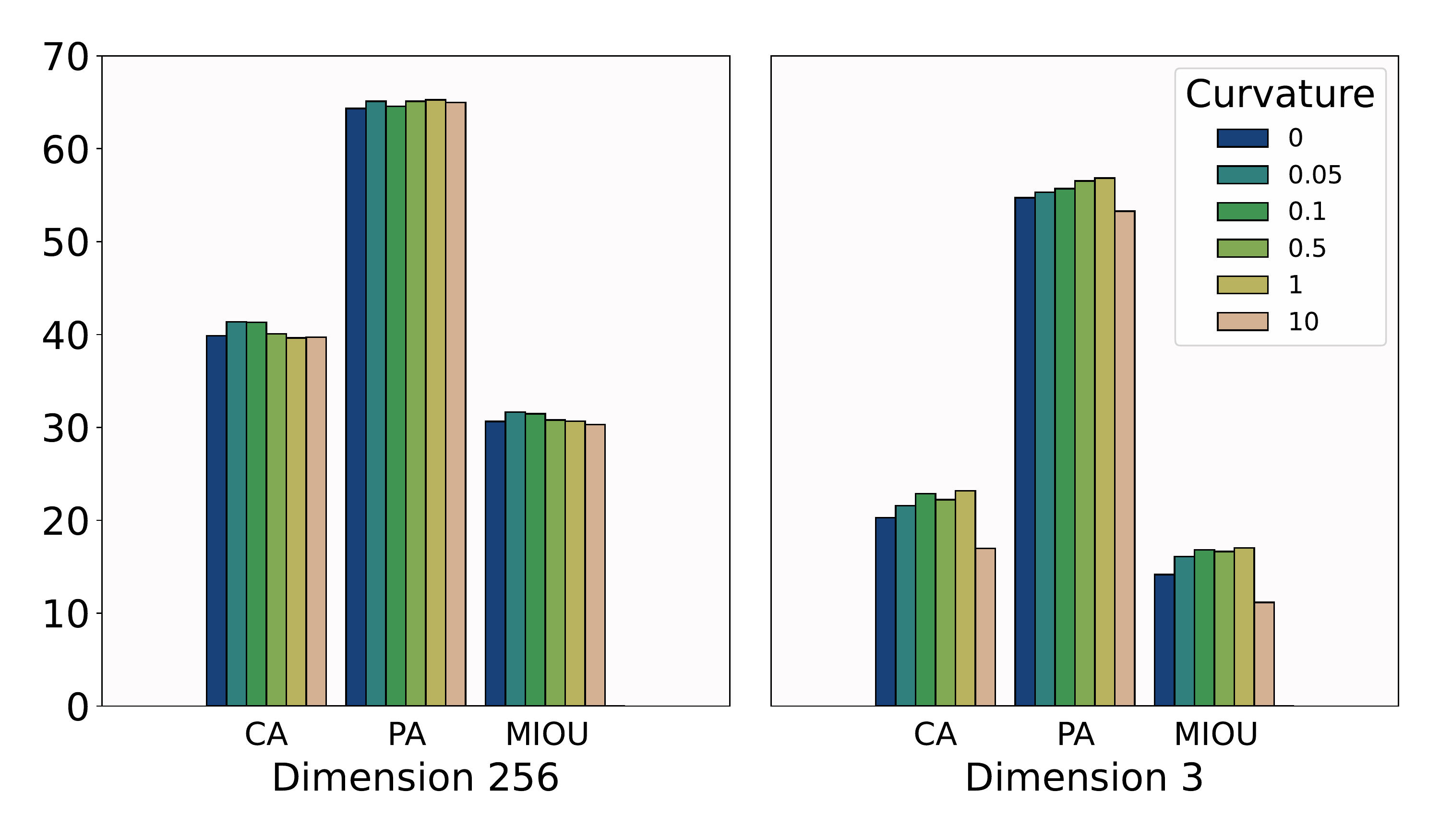}
  \caption{\textbf{Comparison of curvature} for high ($256$) and low ($3$) dimensional hyperbolic embeddings. Performance reported as Classification Accuracy (CA), Pixel Accuracy (PA), and Mean IOU (MIOU). For high dimensions the model is robust to changes in the curvature value, in a low dimensional setting similar robustness can be observed for low curvature values, however the performance drops when using a high curvature value ($10$).}
\label{fig:curv}
\vspace{-0.1cm}
\end{figure}

\textbf{Curvature.} Since hyperbolic spaces are curved, there is an additional hyperparameter compared to the Euclidean space \com{(\ie $c=0$)} that governs the curvature and radius of the Poincar\'e ball. In Figure~\ref{fig:curv} we show the effect of different curvatures
for image segmentation on both 256- and 3-dimensional embeddings. For 256-dimensional embeddings\com{,} we can observe that the \com{effect}
of the curvature value is negligible, with only minor changes in performance even for large curvature difference\com{s} (e.g., $0.05$
to $10$). A similar observation can be made with 3 dimensions, except that for this lower dimensionality we see a drop in performance when the curvature is set to $10$. We suspect that, because the embedding space shrinks with increasing curvature, a low dimensionality combined with a high curvature reduces the size of the embedding space too far. In practice, we use validation to determine the curvature in a range of 0.1 to 2.

\textbf{Hierarchical versus flat hyperbolic softmax.} Throughout the analyses, we have combined hyperbolic embeddings for image segmentation with hierarchical relations amongst the target classes, due to the well-established match between hierarchies and hyperbolic space. In this ablation study, we show the effect of incorporating such hierarchical knowledge in the context of segmentation. We draw a comparison to the conventional flat setting with one-hot encodings over all classes \com{(\ie omitting hierarchies)}. The results shown in Table~\ref{tab:softmax} clearly highlight the benefits of hierarchical softmax, outperforming the flat softmax in almost all cases - on both the hierarchical and the standard metrics. Increasing the dimensionality reduces the difference between the hierarchical and flat softmax, with the flat softmax even slightly outperforming the hierarchical softmax on the standard metric in $256$ dimensions. Nevertheless, across all dimensionalities the hierarchical softmax is preferred for the hierarchical metrics, demonstrating the benefit of incorporating hierarchical knowledge for segmentation.
\begin{table}[t]
\resizebox{1\linewidth}{!}{
\vspace{0.2cm}
\begin{tabular}{l l c ccc}
\toprule
& &  \multicolumn{3}{c}{\textbf{Mean IOU}}\\
Dimension & Softmax & $\sim$ & S & C\\
\midrule
2  & Flat  &  4.31 & 12.47 & 19.48\\
   & Hierarchical  &  8.74 & 22.67 & 33.05\\
   \midrule
3 & Flat  &  11.11 & 26.19 & 34.41\\
  & Hierarchical  &  16.82 & 34.85 & 45.89\\
  \midrule
10  & Flat  &  28.89 & 46.85 & 55.85\\
    & Hierarchical  &  28.99 & 47.35 & 56.74\\
    \midrule
256  & Flat  &  31.77 & 48.59 & 57.27\\
     & Hierarchical  &  31.46 & 48.73 & 58.34\\

\bottomrule
\end{tabular}
}
\caption{\textbf{Effect of embedding hierarchical knowledge} in Hyperbolic Image Segmentation on COCO-Stuff-10k. In few dimensions, employing a hierarchical softmax is preferred over a flat softmax based on one-hot vectors. As dimensionality increases, this preference diminishes for standard metrics, while hierarchical softmax remains preferred for the hierarchical metrics.}
\vspace{-0.35cm}
\label{tab:softmax}
\end{table}
\section{Conclusions}
This work investigates semantic image segmentation from a hyperbolic perspective. Hyperbolic embeddings have recently shown to be effective for various machine learning tasks and data types, from trees and graphs to images and videos. Current hyperbolic approaches do however not scale to the pixel level, since the corresponding operations are memory-wise intractable. We introduce Hyperbolic Image Segmentation, \com{the} first approach for image segmentation in hyperbolic embedding spaces. We outline an equivalent and tractable formulation of hyperbolic multinomial logistic regression to enable this step. Through several analyses, we demonstrate that operating in hyperbolic embedding spaces brings new possibilities to image segmentation, includ\com{ing} uncertainty and boundary information for free, improved zero-label generalization, and better performance in low-dimensional embedding spaces.

\textbf{Limitations and negative impact.}
Throughout the experiments, we have used DeepLabv3+ as backbone due to the well-known and performant nature of the architecture. Our analyses do not yet uncover the effect of hyperbolic embeddings in more shallow or deeper architectures, or their effect beyond natural images such as the medical domain. While we do not focus on specific applications, segmentation in general does have potentially negative societal applications that the reader needs to be aware of, such as segmentation in surveillance and military settings.

\textbf{Acknowledgments.}
Erman Acar is generously funded by the Hybrid Intelligence Project which is financed by the Dutch Ministry of Education, Culture and Science with project number 024.004.022.

{\small
\bibliographystyle{ieee_fullname}
\bibliography{egbib}
}

\end{document}